\documentclass[preprint,11pt]{elsarticle}

\usepackage{amssymb}

\usepackage{url}
\usepackage[]{graphicx}
\usepackage[]{geometry}
\usepackage{amsmath}
\usepackage{setspace}

\usepackage{algorithm}
\usepackage{algorithmic}

\usepackage{url}

\usepackage{multirow}

\usepackage{subfigure}

\usepackage{lscape}

\biboptions{}

\begin{document}

\begin{frontmatter}

\title{Semi-Automatic Construction of a Domain Ontology for Wind Energy Using Wikipedia Articles}

\author[tbtk]{Dilek K\"u\c{c}\"uk\corref{cor1}}
\ead{dilek.kucuk@tubitak.gov.tr}
\author[odtu]{Yusuf Arslan}
\ead{yusuf.arslan@ceng.metu.edu.tr}

\address[tbtk]{Power Electronics Department, T\"UB\.ITAK Energy Institute, 06531, Ankara, Turkey}
\address[odtu]{Department of Computer Engineering, Middle East Technical University, 06531, Ankara, Turkey}
\cortext[cor1]{Corresponding author. Tel: +90-312-2101310; fax: +90-312-2101315}

\begin{abstract}
Domain ontologies are important information sources for knowledge-based systems. Yet, building domain ontologies from scratch is known to be a very labor-intensive process. In this study, we present our semi-automatic approach to building an ontology for the domain of wind energy which is an important type of renewable energy with a growing share in electricity generation all over the world. Related Wikipedia articles are first processed in an automated manner to determine the basic concepts of the domain together with their properties and next the concepts, properties, and relationships are organized to arrive at the ultimate ontology. We also provide pointers to other engineering ontologies which could be utilized together with the proposed wind energy ontology in addition to its prospective application areas. The current study is significant as, to the best of our knowledge, it proposes the first considerably wide-coverage ontology for the wind energy domain and the ontology is built through a semi-automatic process which makes use of the related Web resources, thereby reducing the overall cost of the ontology building process.
\end{abstract}

\begin{keyword}
wind energy\sep wind power\sep domain ontology\sep text mining\sep Web mining
\end{keyword}

\end{frontmatter}

\section{Introduction}\label{sec:intro}
Renewable energy resources (such as wind, solar, and biomass, among others) have emerged as an important engineering topic mainly due to the profound advantages of the corresponding plants over conventional plants (such as fossil-fuelled and nuclear). Considering wind energy, the primary benefits of the wind energy plants are that (i) energy production does not lead to environmental pollution and (ii) as wind is abundant in many countries, the financial costs of these plants are comparably lower than that of the conventional plants \cite{Jain2011}. Compared to the other renewable resources, wind is considerably variable and hence it is hard to determine optimal places for wind power plant installations and to forecast daily energy productions of the plants precisely. The latter point (the uncertainty of the production) results in particular problems during the integration of these plants to the electrical grid as the electricity transmission system operators should be informed of prospective daily productions in advance to make short-term planning and to ensure that the supply matches the demand \cite{Jain2011}. Other directions of research on the topic include predictions for the economic viability of wind power plant projects and wind turbine design \cite{Burton2001}. As wind energy keeps increasing its share in the electricity generation (and annually growing at a rate of 30\% \cite{Wiki-renewables}), research projects targeting at these problems have started to increase (such as \cite{Ritm}).

In this paper, we target at wind power applications from a knowledge-based perspective and propose a wide-coverage domain ontology for wind energy domain\footnote{In this study, we utilize the terms \emph{wind power} and \emph{wind energy} interchangeably to denote the electrical power generated from wind through the wind turbines installed on the wind power plants (as they are commonly used interchangeably and they denote wind generated power), although the terms \emph{power} and \emph{energy} are quite distinct terms in physics.}. A domain ontology is defined as a reusable vocabulary of concepts, relationships, and activities in the domain along with the governing theories and principles \cite{Perez2004}. Various domain ontologies have been reported in the literature \cite{Stevens2002, Gasevic2006, Morbach2007, PQONT, Zhang2011} and there is ongoing research on the topic along with research on the Semantic Web. Although domain ontologies serve various purposes, building them from scratch is too labor-intensive and time-consuming. Hence, we employ a semi-automatic approach to build our wind energy ontology which comprises two main phases: in the first phase, Wikipedia \cite{Wiki-wiki} articles on the topic are processed to learn the concepts from highly frequent words and phrases (n-grams) and in the second phase, these concepts are organized with their properties and relationships to arrive at the ultimate ontology. Yet, the final form of the ontology is open to extensions as well, that is, other concepts and properties covering more details about the domain can be added to the ontology. It can also be integrated with other related engineering ontologies for better coverage of the domain.

Ontology-based studies on wind power domain are rather limited. We come across only three studies on the topic, the first of which \cite{Zhu2008} proposes an ontology on wind power plant information and the resulting ontology comprises some generic concepts which are not specific to wind power domain and moreover it basically aims to cover concepts related to the management of wind power plant information. The second study \cite{Papadopoulos2009} describes an ontology for wind turbines' condition monitoring and is more domain-specific compared to the ontology proposed in \cite{Zhu2008}. Yet this study \cite{Papadopoulos2009} also suffers from low coverage, i.e., it only considers the semantics of wind turbines instead of the whole wind power domain. The last related study \cite{Pipattanasomporn2012} utilizes an ontology of facts to be utilized by a multi-agent system which aims to control a photovoltaic (PV)-based microgrid. This ontology covers specific facts or concepts related to agent operations which are far from representing the domain of renewable energy or wind energy. Hence, to the best of our knowledge, in the current study, we propose the first large-scale ontology for the wind power domain. Yet, as will be emphasized in Section \ref{sec:integration}, this ontology can well be integrated with related engineering ontologies and aligned with the aforementioned ontologies including \cite{Zhu2008, Papadopoulos2009}.

The rest of the paper is organized as follows: In Section \ref{sec:approach}, the ontology building process for wind energy domain is presented in details. Section \ref{sec:integration} is devoted to prospective integration opportunities with related engineering ontologies and plausible application areas of the proposed ontology. Finally, Section \ref{sec:conc} concludes the paper and provides pointers to future work.

\section{Building an Ontology for Wind Energy}\label{sec:approach}
Our semi-automatic ontology construction process for wind energy domain comprises two main phases:
\begin{itemize}
    \item   Learning the concepts and properties of these concepts from related Wikipedia articles by extracting the frequent phrases as unigrams, bigrams, and trigrams from the article texts. This phase executes in fully automated mode.
    \item   Building the ontology by organizing these components and properties utilizing other written information sources like related international standards and textbooks.
\end{itemize}

\begin{figure}[h!]
\center \scalebox{0.81}
{\includegraphics{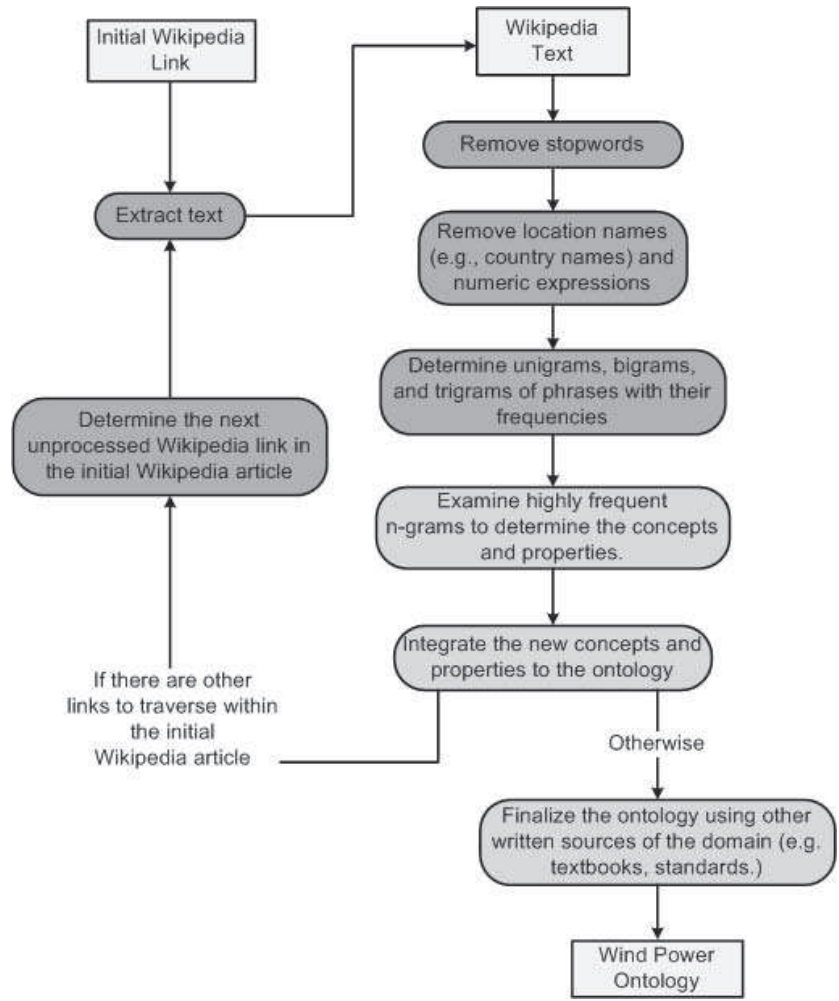}}\caption{Execution Flow of the Employed Ontology Building Approach.}\label{fig:Flow}
\end{figure}

We provide the execution flow of the overall approach with more detailed steps in Figure \ref{fig:Flow}. In the figure, the resources (input and output) are shown as rectangular shapes, while the functional steps are presented as round rectangles (fully automated steps are shown with darker color than the manual steps). At the end of this execution flow, we realize the ontology on Prot\'eg\'e \cite{Protege} which is an ontology editor and a knowledge-base framework. The details of the aforementioned two phases are described in the rest of this section basically following the functional steps in Figure \ref{fig:Flow}.

\subsection{Learning Ontology Concepts from the Web}\label{subsec:learning}
Ontology learning is defined as the task of automatic or semi-automatic knowledge discovery from different data sources and representation of the ultimate ontology \cite{Drumond2008}. Within the course of the current study, we consider Wikipedia \cite{Wiki-wiki} articles as our data source for learning an ontology for the wind energy domain. Wikipedia is a large and valuable source of semantic information. It has been commonly pointed
out in the literature that it can serve various tasks including information extraction, information retrieval, question answering systems, and ontology building \cite{Medelyan2009, Ferrandez2009}. Therefore, we choose the Wikipedia article for the term \emph{wind power} \cite{Wiki-wind-power} as our initial point and utilized this article and linked articles as the information source to facilitate the ontology building process for wind energy domain. The approach followed is very much like that of the Web search engines which traverse the links in Web pages in a breadth-first
manner, yet, we only consider the articles linked to the initial article (links in the first level), i.e., we do not process the articles linked to the linked articles of the initial article.

\begin{table}[h]
  \caption{Frequent N-grams in the Initial Wikipedia Article for Wind Power (with Frequencies in Parentheses).}
  \label{tab:ngrams}
    \centering
    \begin{tabular}{|l|l|l|}
    \hline
    {\it Unigrams} & {\it Bigrams} & {\it Trigrams} \\
    \hline
          wind (346) & wind  power (82) & offshore wind power (8) \\

         power (149) & wind  energy (51) & wind power capacity (4) \\

        energy (135) & wind  turbine (45) & wind power industry (3) \\

       turbine (65) & wind  farm (39) & meteorology and climatology (2) \\

      capacity (54) & renewable  energy (22) & wind turbine engineering (2) \\

    electricity (50) & capacity  factor (16) & electric power transmission (2) \\

          grid (36) & offshore  wind (15) & wind generated power (2) \\
    \hline
    \end{tabular}
\end{table}

Our overall ontology learning procedure for wind energy domain from Wikipedia articles is carried out as follows: first, the text of the initial Wikipedia article is obtained and stopwords in the text are eliminated. After stopword elimination, named entities (such as person, location, and organization names) as well as numeric expressions are also removed from the text. Next, unigrams, bigrams, and trigrams are obtained with their frequencies from the resulting text. Especially highly frequent phrases are then manually examined to determine whether they are relevant to the target ontology or not. Relevant phrases are utilized as concepts and concept properties of the ontology. The same procedure is employed for the
linked Wikipedia articles and again relevant phrases contribute to the ontology. To better illustrate the process, the highly frequent sample phrases in the text of the initial Wikipedia article are provided in Table \ref{tab:ngrams} together with their frequencies in parentheses as sorted by their frequencies.

\subsection{Organizing the Ontology}\label{subsec:organizing}
\begin{landscape}
\begin{figure}[h!]
\center \scalebox{0.57}
{\includegraphics{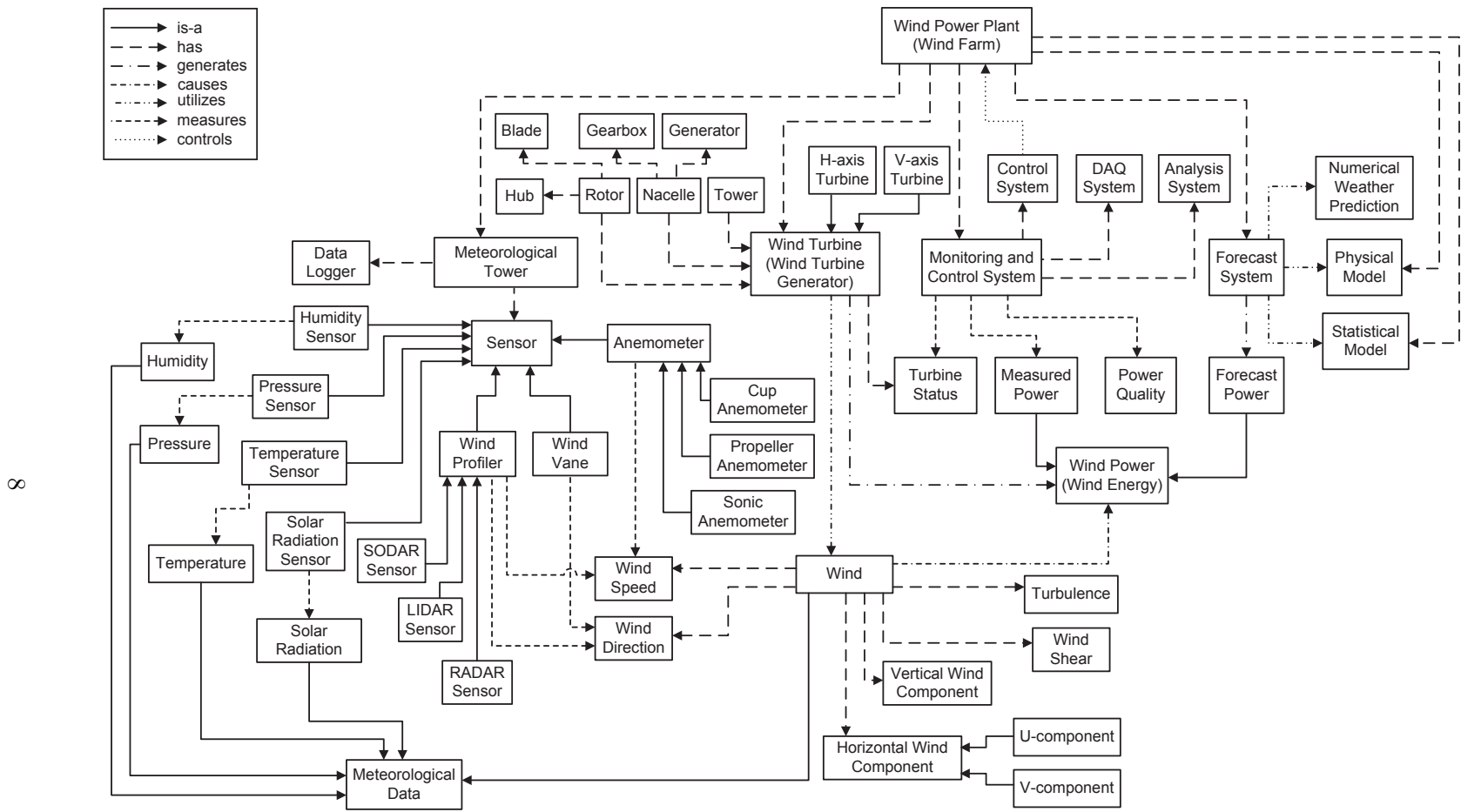}}\caption{The Domain Ontology for Wind Energy.}\label{fig:Ontology}
\end{figure}
\end{landscape}

\begin{figure}[h!]
\center \scalebox{0.55}
{\includegraphics{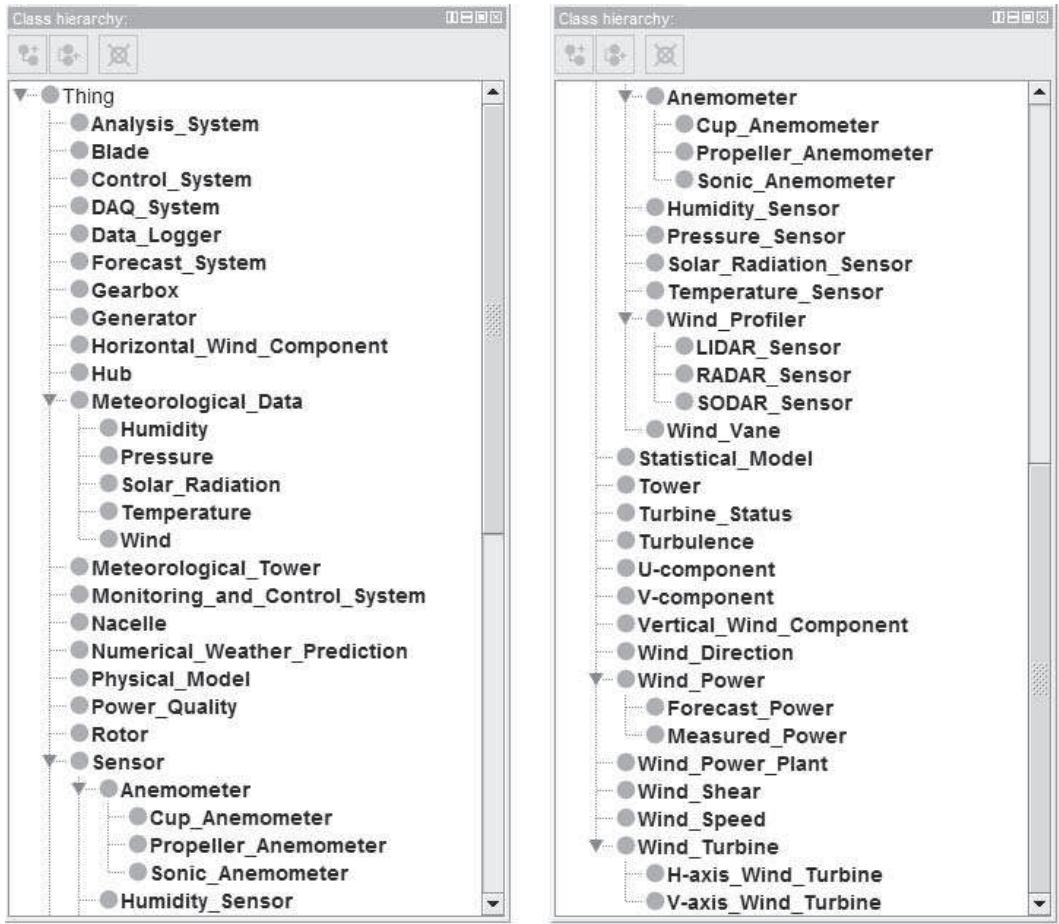}}\caption{The Class Hierarchy of the Wind Energy Ontology on Prot\'eg\'e \cite{Protege}.}\label{fig:Protege1}
\end{figure}

Apart from the frequent phrases determined in the steps described in the previous section, we have also examined several related textbooks of the domain \cite{Burton2001, Jain2011} and the set of wind turbine standards by International Electrotechnical Commission (IEC) \cite{Wiki-iec} to arrive at a wide-coverage (yet extensible) domain ontology for wind power domain. At this stage, the relevant phrases extracted from the Web are complemented with those extracted from the aforementioned written resources and the resulting information is organized to form the ultimate ontology. The taxonomy of the concepts in the final form of the proposed domain ontology is presented in Figure \ref{fig:Ontology}. The ontology is built utilizing Prot\'eg\'e \cite{Protege, Noy2001} as the ontology editor and the hierarchy of the ontology classes is presented in Figure \ref{fig:Protege1} as snapshots.

At the center of the ontology is the \emph{Wind Power Plant} class which has the major components represented by the classes of \emph{Meteorological Tower}, \emph{Wind Turbine}, \emph{Monitoring and Control System}, and \emph{Forecast System}. The \emph{Meteorological Tower} class in turn comprises \emph{Data Logger} and \emph{Sensor} classes, where the latter class has several subclasses, namely, \emph{Humidity Sensor}, \emph{Pressure Sensor}, \emph{Temperature Sensor}, \emph{Solar Radiation Sensor}, \emph{Wind Profiler}, \emph{Wind Vane}, and \emph{Anemometer}. Among these classes, the latter three of them represent devices for measuring wind properties and are therefore more significant for the overall wind energy ontology. To clarify the corresponding devices, a wind profiler is a device for measuring wind speed and direction, while wind vanes only measure wind direction and anemometers only measure wind speed. Anemometers usually come in three basic types represented in our ontology by the classes of \emph{Cup Anemometer}, \emph{Propeller Anemometer}, and \emph{Sonic Anemometer}. All of the \emph{Sensor} subclasses in the proposed wind energy ontology represent devices for measuring distinct types of data on meteorological phenomena which are represented by the \emph{Meteorological Data} class. The \emph{Wind} class has the components of \emph{Wind Speed}, \emph{Wind Direction}, \emph{Vertical Wind Component}, \emph{Horizontal Wind Component} (which in turn has \emph{ U-component} and \emph{V-component} as components), \emph{Wind Shear} and \emph{Turbulence}. The names of the first four of these classes are fairly self-explanatory. Considering the last two classes, wind shear is the change of wind speed as a function of height while turbulence is defined to be the standard deviation of horizontal wind speed, vertical wind speed, and wind direction as 10-min averages \cite{Jain2011}.

The second significant component of a wind farm is a set of wind turbines (i.e., wind turbine generators) represented with the \emph{Wind Turbine} class. The turbines usually fall into one of the two types: horizontal axis turbines (modeled with \emph{H-axis Turbine}) and vertical axis turbines (modeled with \emph{V-axis Turbine} class). Main components of a generic wind turbine are rotor, nacelle, and tower which are modeled with the corresponding classes of \emph{Rotor}, \emph{Nacelle}, and \emph{Tower}, respectively. The rotor of a wind turbine is usually composed of the blades (modeled by the \emph{Blade} class) and the rotor hub (represented with the \emph{Hub} class) which connects the blades to the shaft \cite{Jain2011}. Nacelle is the part of the turbine that includes the main parts apart from those included in the rotor system such gearbox, generator, and brake \cite{Jain2011} where we only include the \emph{Gearbox} and the \emph{Generator} classes in our ontology as parts of the \emph{Nacelle} class. Depending on the requirements of the application in which the wind energy ontology will be utilized, the parts of the \emph{Nacelle} class can be extended and more classes to represent these parts can be added to the ontology. The last part of a wind turbine is the structural part represented with the \emph{Tower} class. Wind turbines generate wind power represented with the \emph{Wind Power} class in our ontology.

The last two parts of a typical wind power plant are the monitoring and control system (modeled with \emph{Monitoring and Control System} class) and the forecast system (modeled with \emph{Forecast System} class). The former system has several components including a control system, a data acquisition (DAQ) system, and an analysis system denoted by the classes of \emph{Control System}, \emph{DAQ System}, and \emph{Analysis System}, respectively. This system is used to monitor and measure turbine status information, as well as generated power and electrical power quality, which are represented by \emph{Turbine Status}, \emph{Measured Power}, and \emph{Power Quality} classes, respectively. The power quality is actually another related domain which is not modeled in our ontology in details but as will be emphasized in the following section, an ontology for this domain can be integrated into the current ontology to cover this power quality concept and related concepts adequately. The \emph{Forecast System} class represents the system which utilizes numerical weather predictions together with previously generated physical and statistical models of the wind power plants to produce forecasts regarding the wind generated power. Hence, related classes with self-explanatory names include \emph{Numerical Weather Prediction}, \emph{Physical Model}, \emph{Statistical Model}, and finally \emph{Forecast Power}. Although some forecast systems may also output forecasts for wind speed and direction, we do not consider that case here as the ultimate aim of the system is usually power forecast.

\emph{Wind Power Plant}, which is the central class of our wind energy ontology, has the synonymous names of \emph{Wind Farm}, \emph{Wind Plant}, and \emph{Wind Energy Plant}. It is also commonly abbreviated as \emph{WPP}. As other concepts of the domain also have several synonyms, we have added a \emph{synonymSet} property for applicable concepts to hold the synonyms of the actual class names corresponding to these concepts. Abbreviations are also treated as synonyms and this property is used to hold abbreviations as well. We provide synonyms of some of the concepts of the ontology in Table \ref{tab:synsets} for illustrative purposes. As shown in the table, \emph{WPP} is added as a synonym of the \emph{Wind Power Plant} class. Similarly, \emph{WTG} is the abbreviation for \emph{Wind Turbine Generator} which in turn is a synonym of the \emph{Wind Turbine} class, hence \emph{WTG} is also added to the synonym set of this class.

\begin{table}[h]
  \caption{Synonyms of Some of the Concepts in the Wind Energy Ontology.}
  \label{tab:synsets}
    \centering
    \begin{tabular}{|l|c|}
    \hline
    \emph{Ontology Class} & \emph{Synonyms} \\
    \hline
    Wind Power Plant & WPP, Wind Plant, Wind Energy Plant, Wind Farm \\
    \hline
    & Wind Power Generation, Energy, Generation, \\
    Wind Power & Wind Energy, Wind Generation, Wind Generated Power, \\
    & Electricity Production \\
    \hline
    Wind Turbine & Turbine, Wind Turbine Generator, WTG, Generator \\
    \hline
    Wind Vane & Weather Vane, Weather Cock \\
    \hline
    \end{tabular}
\end{table}

As we have previously pointed out, properties and relations are also specified within the ultimate wind energy ontology. While properties are mostly obtained from the extracted keywords from Wikipedia articles (where the procedure is outlined in the previous subsection), relations are manually formed considering the interrelations between the semantic concepts (corresponding to the classes in Figure \ref{fig:Ontology}). Plausible properties of some of the classes are provided in Table \ref{tab:properties} for illustrative purposes. It should be noted that properties may also have some synonymous names. For instance, \emph{installed capacity} property of \emph{Wind Power Plant} class has \emph{rated capacity}, \emph{nominal capacity}, \emph{maximum effect}, \emph{power capacity}, \emph{nameplate capacity}, and \emph{wind power capacity} as its synonyms. Similarly, \emph{utilisation rate} is a synonym for the \emph{capacity factor} property of the same class\footnote{\emph{Installed capacity} of a power plant is defined as the ``intended technical full{–}load sustained output" \cite{Wiki-wind-power} of the corresponding plant while \emph{capacity factor} is usually defined as the ratio of generated power to the \emph{installed capacity}.}. Therefore such information is also modeled in the ontology through the \emph{synonymSet} property of the applicable concept properties. Among the properties of \emph{Wind Power Plant} class in Table \ref{tab:properties}, \emph{voltage level} corresponds to that level through which the plant is integrated to the electrical grid and the names of the remaining ones are fairly self-explanatory. The properties of \emph{Wind Turbine} mostly model the technical characteristics of the turbines while the properties of the \emph{Wind Speed} class aim to cover the actual speed, the height at which the speed is measured or forecasted, the date of measurement or forecast, and the plant where the speed is measured or forecasted.

\begin{table}[h]
  \caption{Properties of Some of the Concepts of the Wind Energy Ontology.}
  \label{tab:properties}
    \centering
    \begin{tabular}{|l|c|}
    \hline
    \emph{Ontology Class} & \emph{Properties} \\
    \hline
    & name, owner, location, license date, \\
    Wind Power Plant & number of turbines, installed capacity, capacity factor,\\
    & voltage level\\
    \hline
    & model, hub height, swept area, rated power,\\
    Wind Turbine & power curve, cut-in wind speed, cut-out wind speed,\\
    & rotor diameter, number of blades\\
    \hline
    Wind Speed & speed, height, date, wind power plant\\
    \hline
    \end{tabular}
\end{table}

Considering the relations among the concepts in the ontology, there are basically seven distinct relation types which are outlined below:
\begin{itemize}
    \item   \emph{is-a}: This type denotes the subclass-class relation between the corresponding concepts in the ontology. For instance, \emph{is-a} relation exists between the \emph{Anemometer} and \emph{Sensor} classes to specify that an anemometer is a kind of sensor.
    \item   \emph{has}: This relation corresponds to the part-of information between concepts. It exists between the \emph{Wind Power Plant} and \emph{Wind Turbine} classes as wind power plants have a number of turbines.
    \item   \emph{generates}: This relation type and the remaining types denote functional relations with quite self-explanatory names. Hence, we only provide examples from the ontology in Figure \ref{fig:Ontology} to illustrate the relations, hence, this \emph{generates} relation can be observed between \emph{Wind Turbine} and \emph{Wind Power} classes of the ontology.
    \item   \emph{causes}: This type of relation is illustrated between \emph{Wind} and \emph{Wind Power} classes as depicted in Figure \ref{fig:Ontology}.
    \item   \emph{utilizes}: This relation exists between the classes of \emph{Wind Turbine} and \emph{Wind} as wind turbines utilize wind to generate wind power.
    \item   \emph{measures}: This relation type can be observed between the \emph{Anemometer} and \emph{Wind Speed} classes in the ontology.
    \item   \emph{controls}: This relation exists between \emph{Control System} and \emph{Wind Power Plant} classes in the wind energy ontology.
\end{itemize}

It should be noted that we do not claim to cover all meteorological phenomena and other wind related concepts in our ontology and as it will be clarified in the upcoming section, relevant ontologies on the topic can be integrated into the wind power ontology to increase its coverage.

The final form of the proposed ontology is made available for research purposes at \url{http://www.ceng.metu.edu.tr/~e120329/wont.owl} as a Web Ontology Language (OWL) file.

\section{Prospective Integration with Related Ontologies and Application Areas}\label{sec:integration}
The proposed wind energy ontology has some concepts, particularly \emph{Power Quality} and \emph{Meteorological Data}, corresponding to distinct but related domains which may be represented in a more detailed manner. But, instead of extending the proposed wind energy ontology with the concepts for these domains from scratch, ontologies for these two domains can be integrated into the wind energy ontology. For the domain of meteorological data, the ontology covered in \cite{Ramachandran2006}, which is an ontology for atmospheric science, can be integrated into the wind energy ontology. For the domain of electrical power quality, PQONT domain ontology \cite{PQONT} has been proposed and it can readily be integrated into the wind power ontology as it is publicly available as an OWL file. Another plausible option is to extend the proposed ontology to make it a larger renewable energy ontology, considering the semantics of other renewable energy types like solar and biomass. With this prospective extension and aforementioned integration opportunities with the related domain ontologies, the wind energy ontology can be turned into a wider coverage ontology and hence can serve a wider range of applications.

There are several significant application areas of domain ontologies. Basically, they are known to be of particular importance for semantic information extraction, information retrieval, and question answering systems, among others. Below, we outline the plausible application areas in which the wind energy ontology that we have proposed can be employed:
\begin{itemize}
    \item   The ontology could act as a shared vocabulary of concepts to ensure the interoperability of the various wind energy applications like monitoring and forecast systems. The proposed ontology could also aid in reducing the software analysis and design costs during the development of the aforementioned systems as it acts as a viable source of semantic information for the underlying domain of wind energy.
    \item   The ontology can be integrated into a domain-specific information extraction system to automatically extract significant pieces of information on wind energy and related topics from free natural language texts. Utilizing such an information extraction system, the proposed ontology could be populated with relevant instances. Similar to the previous point, the ontology could be employed as part of an information retrieval system to determine the index terms to be used during retrieval or as part of a question answering system to determine relevant answers from relevant documents for the specified questions. A text categorization system can also benefit from the proposed ontology to detect documents related to wind energy (or, renewable energy) in a large set of text documents. The ontology could also be employed in a natural language interface over the populated ontology where queries in natural language can be transformed into valid ontology queries making use of the concept names, properties, and their synonyms to pinpoint the queried information. This final application is exemplified in \cite{PQONT} where a natural language interface, based on the PQONT domain ontology for electrical power quality, is described where the interface basically utilizes the ontology to determine the concepts specified in the natural language query expressions..
\end{itemize}

\section{Conclusion}\label{sec:conc}
There is a growing interest in wind energy which is a significant type of renewable energy with a growing share in the electricity production all over the world. Accordingly, research on topics related to wind energy, including monitoring and forecast of wind power, has gained considerable attention. In this study, we deal with the wind energy domain from a knowledge-based perspective and propose the first considerably wide-coverage domain ontology for wind energy. The ontology building process is a learning-based semi-automated procedure where in the first phase plausible keywords to be utilized during the determination of the concepts and their properties are learned from the Web. Namely, related Wikipedia articles are processed to determine the relevant keywords and next related textbooks and international standards are examined to organize these keywords to arrive at the ultimate ontology. Hence, we decrease the ontology building costs considerably by employing a learning procedure and utilizing Web as a source of semantic information. It should be noted that the proposed ontology can be extended or customized to better satisfy the needs of prospective applications in which the ontology will be employed. The final form of the proposed ontology is also made publicly available for research purposes.

Plausible directions of future research include the integration of the ontology with related engineering ontologies such as those for atmospheric science and electrical power quality. It can also be extended into a more comprehensive renewable energy ontology. Moreover, the ontology can be employed in knowledge-based systems such as information extractors, information retrieval, question answering, and text categorization systems in addition to natural language interfaces, to assess its practical contribution to such systems.

\bibliographystyle{elsarticle-num}
\bibliography{ontology}

\begin{thebibliography}{10}
\expandafter\ifx\csname url\endcsname\relax
  \def\url#1{\texttt{#1}}\fi
\expandafter\ifx\csname urlprefix\endcsname\relax\def\urlprefix{URL }\fi
\expandafter\ifx\csname href\endcsname\relax
  \def\href#1#2{#2} \def\path#1{#1}\fi

\bibitem{Jain2011}
P.~Jain, Wind Energy Engineering, 1st Edition, McGraw-Hill, 2011.

\bibitem{Burton2001}
T.~Burton, D.~Sharpe, N.~Jenkins, E.~Bossanyi, Wind Energy Handbook, 1st
  Edition, John Wiley \& Sons, 2001.

\bibitem{Wiki-renewables}
{Renewable energy - {Wikipedia}},
  \url{http://en.wikipedia.org/wiki/Renewable_energy}.

\bibitem{Ritm}
{Wind Power Monitoring and Forecast System (R\.ITM)},
  \url{http://www.ritm.gov.tr/en/root/index.php}.

\bibitem{Perez2004}
A.~G\'omez-P\'erez, M.~Fern\'andez-L\'opez, O.~Corcho, Ontological Engineering,
  3rd Edition, Springer-Verlag, 2004.

\bibitem{Stevens2002}
R.~Stevens, C.~Goble, I.~Horrocks, S.~Bechhofer, Building a bioinformatics
  ontology using {OIL}, IEEE Transactions on Information Technology in
  Biomedicine 6~(2) (2002) 135--141.

\bibitem{Gasevic2006}
D.~Gasevi\'c, V.~Devedzi\'c, Petri net ontology, Knowledge-Based Systems 19~(4)
  (2006) 220--234.

\bibitem{Morbach2007}
J.~Morbach, A.~Yang, W.~Marquardt, Ontocape{—-}a large-scale ontology for
  chemical process engineering, Engineering Applications of Artificial
  Intelligence 20~(2) (2007) 147--161.

\bibitem{PQONT}
D.~K\"{u}\c{c}\"{u}k, O.~Salor, T.~\.{I}nan, I.~\c{C}ad{\i}rc{\i},
  M.~Ermi\c{s}, {PQONT}: A domain ontology for electrical power quality,
  Advanced Engineering Informatics 24 (2010) 84--95.

\bibitem{Zhang2011}
C.~Zhang, C.~Cao, Y.~Sui, X.~Wu, A chinese time ontology for the {Semantic
  Web}, Knowledge-Based Systems 24~(7) (2011) 1057--1074.

\bibitem{Wiki-wiki}
{Wiki - {Wikipedia}}, \url{http://en.wikipedia.org/wiki/Wiki}.

\bibitem{Zhu2008}
Y.~Zhu, X.~Wang, D.~Cheng, Ontology-based research on wind power plant
  information interaction, in: Proceedings of the Workshop on Knowledge
  Discovery and Data Mining, 2008.

\bibitem{Papadopoulos2009}
P.~Papadopoulos, L.~Cipcigan, Wind turbines' condition monitoring: an ontology
  model, in: Proceedings of the International Conference on Sustainable Power
  Generation and Supply, 2009.

\bibitem{Pipattanasomporn2012}
M.~Pipattanasomporn, H.~Feroze, S.~Rahman, Securing critical loads in a
  pv-based microgrid with a multi-agent system, Renewable Energy 39~(1) (2012)
  166--174.

\bibitem{Protege}
{The {Prot\'eg\'e} {Ontology Editor and Knowledge Acquisition System}},
  \url{http://protege.stanford.edu}.

\bibitem{Drumond2008}
L.~Drumond, R.~Girardi, A survey of ontology learning procedures, in:
  Proceedings of the 3rd Workshop on Ontologies and Their Applications, 2008.

\bibitem{Medelyan2009}
O.~Medelyan, D.~Milne, C.~Legg, I.~H. Witten, Mining meaning from {Wikipedia},
  International Journal of Human-Computer Studies 67~(9) (2009) 716--754.

\bibitem{Ferrandez2009}
S.~Ferr\'andez, A.~Toral, O.~Ferr\'andez, A.~Ferr\'andez, R.~Munoz, Exploiting
  {Wikipedia} and {EuroWordNet} to solve cross-lingual question answering,
  Information Sciences 179~(20) (2009) 3473--3488.

\bibitem{Wiki-wind-power}
{Wind power - {Wikipedia}}, \url{http://en.wikipedia.org/wiki/Wind_power}.

\bibitem{Wiki-iec}
{{IEC} 61400 - {Wikipedia}}, \url{http://en.wikipedia.org/wiki/IEC_61400}.

\bibitem{Noy2001}
N.~F. Noy, M.~Sintek, S.~Decler, M.~Crubezy, R.~W. Fergerson, M.~A. Musen,
  Creating {Semantic Web} contents with {Protege-2000}, IEEE Intelligent
  Systems 16~(2) (2001) 60--71.

\bibitem{Ramachandran2006}
R.~Ramachandran, S.~Movva, S.~Graves, S.~Tanner, Ontology-based semantic search
  tool for atmospheric science, in: Proceedings of the 22nd International
  Conference on Interactive Information Processing Systems for Meteorology,
  Oceanography, and Hydrology, 2006.

\end{thebibliography}

\end{document}